# Multi-view deep learning for reliable post-disaster damage classification

ASIM B. KHAJWAL, CHIH-SHEN CHENG
and ARASH NOSHADRAVAN

**ABSTRACT**

This study aims to enable more reliable automated post-disaster building damage classification using artificial intelligence (AI) and multi-view imagery. The current practices and research efforts in adopting AI for post-disaster damage assessment are generally (a) qualitative, lacking refined classification of building damage levels based on standard damage scales, and (b) trained based on aerial or satellite imagery with limited views, which, although indicative, are not completely descriptive of the damage scale. To enable more accurate and reliable automated quantification of damage levels, the present study proposes the use of more comprehensive visual data in the form of multiple ground and aerial views of the buildings. To have such a spatially-aware damage prediction model, a Multi-view Convolution Neural Network (MV-CNN) architecture is used that combines the information from different views of a damaged building. This spatial 3D context damage information will result in more accurate identification of damages and reliable quantification of damage levels. The proposed model is trained and validated on reconnaissance visual dataset containing expert-labeled, geotagged images of the inspected buildings following hurricane Harvey. The developed model demonstrates reasonably good accuracy in predicting the damage levels and can be used to support more informed and reliable AI-assisted disaster management practices.

## INTRODUCTION

The ability to assess the extent of damage in the aftermath of large-scale disaster events is crucial for effective post-disaster planning, management, and fund-allocations. However, due to vast spatial extent of disaster damage and limited resources, coupled with inefficient methods of damage assessment, the process takes several weeks. This often leads to significant delays in the recovery process as well as extended economic losses [1] [2]. Motivated by overcoming time, accessibility, and safety concerns, new practices have emerged to leverage advances in remote sensing and AI for rapid and automated post-disaster assessment of damaged infrastructure. It is expected these solutions converge with human experts to form human-in-the-loop schemes paving the way for faster and sufficiently accurate damage assessment operations. Specifically, in the domain of AI-assisted disaster damage assessment, several deep-learning (DL) models, particularly Convolutional Neural Network (CNN) applications have been proposed in the literature to perform post-disaster damage assessment at infrastructure level [3] [4] [5] [6] [7]. Xu et al. [8] and Gupta et al. [9] presented CNN-based models that recognizes damaged buildings in satellite images. Chen et al. [10] explored deep vision models for damage evaluation in the aftermath of a tornado event. Pi et al. [11] adopt deep vision models to detect damaged and undamaged roofs in post-hurricane footages. Cheng et al. [12] proposed a stacked CNN architecture for automated preliminary damage assessment in buildings.

---

[1] Zachry Department of Civil and Environmental Engineering, Texas A&M university, College Station, TX, 77843, USA.

While these developments are promising towards realizing scalable and rapid automated disaster assessment, challenges remain to be addressed for the effective use of this new technology in practice. Firstly, most of the studies are restricted to qualitative damage identification in the built infrastructure rather than the refined quantification and classification of damage extent with respect to standard damage scales. Secondly, most of the damage classifications in buildings are based on individual images (usually satellite view), which may be indicative at times but not necessarily descriptive of the exact damage state of the buildings. It is expected that the classification outcomes corresponding to multiple views of the same building be different because of factors such as directional damage, partial visibility, and image quality when the same building is viewed from different angles or heights [12].

In the present research, the aim is to address this gap by constructing a spatially-aware MV-CNN architecture that combines the information from different views of a post-disaster damaged building (aerial view, multiple street views), resulting in 3D aggregation of the 2D damage features from each view. The applicability and advantages of the proposed approach is presented and illustrated using a real-word post-hurricane damage data. The presented MVCNN prediction is found to be more reliable and accurate due to the more informative inputs.

## METHODOLOGY

### Dataset Description

The presented study uses the post-hurricane visual damage data collected in the aftermath of Hurricane Harvey, a category 4 hurricane made its landfall on August 25, 2017, near Rockport, Texas. The post-hurricane data has been collected by expert post-disaster reconnaissance teams from NSF-supported StEER (Structural Extreme Event Reconnaissance) [13] network. The data is in the form of multiple geotagged images of the inspected buildings along with the detailed observations and overall rating based on the on-site assessment and evaluation of the damaged buildings [14]. The visual data collated for the study from the database includes both the geo-tagged ground images taken from different sides, as well as the aerial and satellite photographs of the affected buildings. For uniformity, five different images of each building considered for the study is selected from the database. The selected images consist of four different ground images and a satellite image of each building. Images containing unnecessary information or taken from inside of the building are filtered out and not considered in the study. Figure 1 shows the visual data for a representative building used in the study.

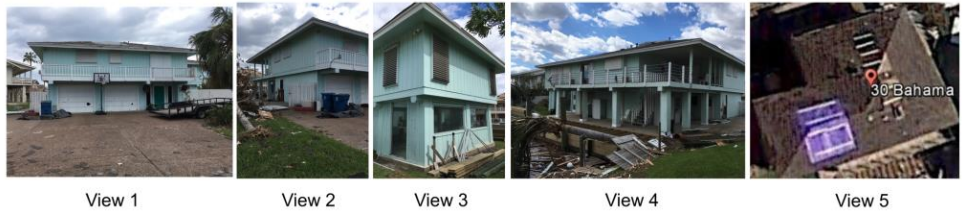
Figure 1. Sample of the visual data compiled for a single building.

Data annotation is performed using Labelbox [15], an online annotation and training data management platform. In total images corresponding to 400 buildings (2000 images) were annotated and assigned labels representing their damage state. The damage labels (0 to 5) were based on the expert assessment reported in the original database. The entire annotated dataset is split into training (80%), validation (10%), and testing sets (10%).

**Model Description**

Contrary to several visual recognition tasks involving detection of everyday objects in natural settings, identification, and classification of post-disaster damage in buildings is a highly complex task. This is mainly because (i) different damage states are not uniquely defined and hence inherently difficult to differentiate, and (ii) Each building is visually different than the other and the damage description in different buildings may not be unique. This can potentially result in relatively large number of incorrect predictions, negatively impacting the model performance. To minimize the erroneous detections, a PDA model with stacked structure (Figure 2) is adopted in this work, which consists of two CNNs. Specifically, the stacked model consists of a *building localization model* (Model-L) to differentiate "buildings" from "no building" objects, and a *damage classification model* (Model-C) to perform the actual task of building damage level assessment. Further details about these models are discussed in subsequent sections.

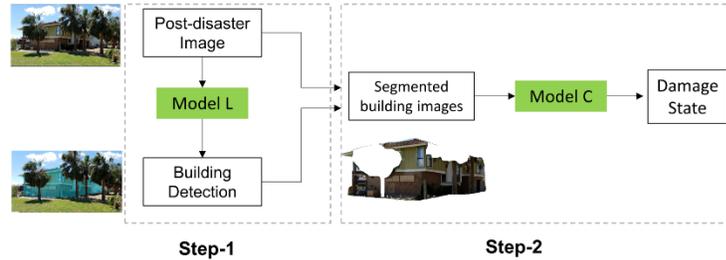
Figure 2. Schematic representation of the proposed stacked PDA.

BUILDING LOCALIZATION MODEL

The first part of the adopted stacked model is a building localization model (Model-L), which is tasked with segmentation or localization of buildings in each image irrespective of the level of damage incurred by the building. The model is trained using pixel level annotations of the buildings performed using the Labelbox platform. The idea is to recognize the pixels belonging to buildings in an image and filter out the visual information unnecessary for determining the damage state of the building like trees, sky, roads, etc. Model-L is developed using transfer learning on a semantic segmentation model architecture called PSPNet [16]. The adopted segmentation model uses ResNet-50 [17] as the backbone for feature extraction, its weights pre-trained on the ImageNet dataset [18].

DAMAGE CLASSIFICATION MODEL

The damage classification model (Model-C) which forms the second part of the adopted stacked model, classifies the segmented buildings generated by Model-L into five damage states based on the level of damage incurred by the buildings in question. The damage states are based on the guidelines in the HAZUS-MH hurricane model residential damage scale [19]. Model-C is thus an approximated function $f: x \rightarrow \{1, ..., k\}$ that maps input building image $x$ to its corresponding damage level $y \in \{1, ..., k\}$. It is based on the proposed composite network called Multi-view Damage CNN (MVD-CNN). The proposed MVD-CNN model is built on feature extraction from 2D images using standard pre-trained CNN architectures. The extracted feature maps from different building views are combined into a common 3D feature space using a multi-view pooling layer. The combined feature space is then passed into a series of fully-connected layers for the final damage level prediction (Figure 3). Mathematically, the underlying fact is that all the building images are organized in terms of a collection of images, $X_V$ where each collection corresponds to each individual building and consists of $n_V$ distinct views of the same building. Hence, the classification task at hand may now be reformulated as:

$$f: X_V \rightarrow \{1, ..., k\} \mid X_V = \{x^{(1)}, x^{(2)}, ..., x^{(n_V)}\} \qquad (1)$$

where $\{1, ..., k\}$ denotes the k possible damage states a building can be classified into, $x^{(i)}$ denotes one image corresponding to a specific view of the building in question. MobileNet [20] is chosen as the base model for the feature extraction for each of the considered views separately.

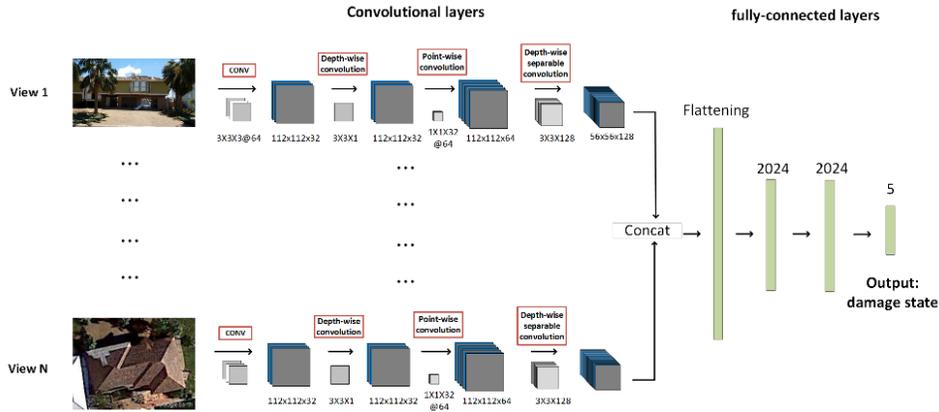

Figure 3. Proposed multi-view classification model (Model-C).

**RESULTS AND DISCUSSION**

**Model Implementation**

TRAINING THE LOCALIZATION MODEL (MODEL-L)

The PSPNet semantic segmentation model, pre-trained on the ImageNet dataset, is retrained on 60% of the compiled dataset and validated on 20% of the data. Data augmentation is employed as a technique to enhance the size of the training data and prevent model overfitting. Different augmentation operation applied to the training dataset include horizontal flip, affine transforms, perspective transforms, brightness/contrast/colors manipulations, image-blurring, and sharpening, Gaussian noise, and random crops. Categorical focal loss is used as the loss function as it works well for highly imbalanced class scenarios [21]. Adam optimizer [22] with a learning rate of 0.0001 and a batch size of 1 is used. The training and validation loss is monitored for 50 epochs and the training is terminated once the validation loss starts to increase, indicating the onset of overfitting.

TRAINING THE CLASSIFICATION MODEL (MODEL-C)

The classification model, Model-C consists of the composite convolution base (MobileNet) with pre-trained weights, and the newly added fusion layer and the fully connected network for final damage state prediction. Corresponding to each of the 5 views considered in the study, feature extraction is performed using transfer learning on a MobileNet CNN model. Following this, the convolutional feature maps obtained from different CNNs corresponding to each view are stacked and subsequently processed together. The details about this fusion technique, referred to as early fusion can be found in [23]. The adopted fusion technique preserves the correspondence between the feature maps of different views. The newly added fully connected layers are initialized with random weights, which are updated through retraining the model on the training dataset. The following methodology is adopted for training the classification model (Model-C):
   a. First, the weights in the pre-trained convolutional layer are frozen and the model is trained for 25 epochs with a learning rate of 0.001. This allows the model to adapt the pre-trained features to the new data.
   b. In the second step, the weights in the convolution base are unfrozen and the entire model is retrained at a smaller learning rate of 0.0001. This is called fine-tuning and allows the incremental improvements and fine modifications of the pre-trained weights to learn more effective features.

In addition to this, data augmentation technique is used to enhance the training dataset. This exposes the network to a more diverse training data, resulting in a more generalizable output.

**Model Testing and Performance Evaluation**

To study the model performance, common CNN performance metrics for object detection or pixel-vise semantic segmentation are calculated. These metrics are Intersection over Union (IoU), precision, recall, and average precision (AP) [24]. IoU, also called Jacard index, measures the percent overlap between the target mask and the prediction output [25]. IoU measures the area common between the target (ground-truth) and the prediction mask as a ratio of the total area across both masks. Other metrics to evaluate the performance of a semantic segmentation model are precision, recall and AP, expressed in terms of the number of true-positive (TP), false-positive (FP), and false-negative (FN) predictions. Precision indicates the number of objects matching the ground truth (TP) as a ratio of the total number of objects predicted in each image (TP+FP), while recall is defined as the ratio of positive predictions (TP) over the number of objects annotated in our ground truth (TP+FN) [26]. When plotted together, the area under the resulting precision-recall curve is termed as AP [24]. To filter out low confidence predictions by Model-L, a threshold IoU score of 0.5 is used. An observation is classed as TP when the prediction-target mask pair has an IoU score exceeding 0.5. The performance metrics observed for our localization Model-L on the testing data are given in Table I. Based on the precision-recall curve shown in Figure 4, the F-1 score, defined as the weighted average of precision and recall, is observed to be 80%. Mean Average Precision (mAP) for the model comes out to be 91.98%.

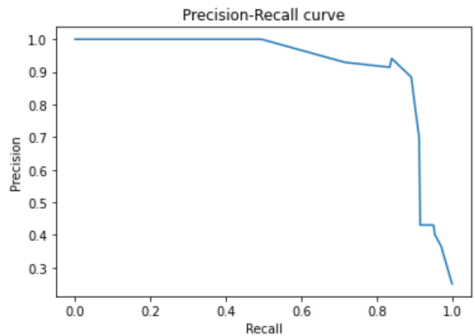

Figure 4. Precision-Recall curve for Model-L

TABLE I. EVALUATION METRICS FOR MODEL-L ON TESTING DATA

|  | Mean IoU | Mean F1 score | Mean Precision | Mean Recall |
| --- | --- | --- | --- | --- |
| Model-L | 76.025 % | 80.093 % | 88.391 % | 89.192 % |

The performance of the multi-class classification Model-C is evaluated in terms of the testing accuracy, representing the percentage of correct predictions by the model. The trained Model-C yields an overall testing accuracy of around 52%. The confusion matrix showing the comparison of actual damage states with the damage states predicted by our trained model is shown in Figure 5. It can be observed that the model identifies the DS-4 accurately, given that the damage characteristics for DS-4 (complete damage) are significantly distinct as compared with other damage states. Also, the model seems to be confused about distinguishing adjacent damage states, for example 70% of DS-0 are classified as DS-1, similarly 58% of DS-2 are classified as DS-1. However, Model-C shows significantly better performance in identifying correct damage state at a coarser level. If we redefine 5 damage states into 3 damage levels, i.e., Minor damage (DS-0 and DS-1), Moderate damage (DS-2 and DS-3) and Extreme damage (DS-4), the testing accuracy increases to 68%. The corresponding confusion matrix (Figure 5) shows significant improvement in predicting the correct damage states.

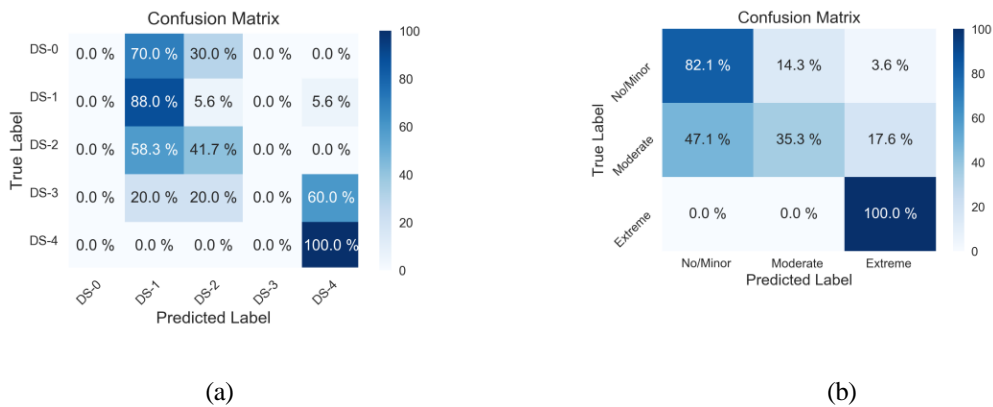

Figure 5. Confusion matrix for Model-C based on: (a) Original 5-level damage classification, (b) modified 3-level damage classification.

## CONCLUSION

The goal of this paper was to develop an AI-assisted post-disaster damage assessment model for reliable and accurate prediction of damage states in residential buildings. Extensive training dataset was compiled from an open-source data repository and the annotated images depicting the damage in the captured buildings was used to train and test the developed deep learning models. To address the limitations and improve the existing practices in the use of artificial intelligence for rapid and automated disaster damage assessment, the paper proposes a multi-view deep learning model that makes predictions based on the combined information from different views of the damaged building. This spatial 3D context damage information will result in more accurate identification of damages and reliable quantification of damage levels.

A stacked PDA model consisting of dual CNN architectures was developed, consisting of (a) a localization model (Model-L) to first localize the build objects in the images, and (b) a multi-view classification model (Model-C) that uses the output from the previous model as input and predicts the damages state of the building in question. Model-L was trained using transfer learning on a pre-trained ResNet-50 backbone, yielding mAP of 91.98%. Model-C was also trained using transfer learning based on MobileNet backbone. An overall testing accuracy of 52% was observed using Model-C when considering classification into 5 different damage states. However, redefining the damage classification at a coarser level (having 3 damage states – Minor, Moderate, and Extreme damage) resulted in better performance by the trained model, enhancing the testing accuracy to 68%.